%% file: acl2021.tex
\documentclass[11pt,a4paper]{article}
\usepackage[hyperref]{acl2021}
\usepackage{times}
\usepackage{latexsym}
\pdfoutput=1

\aclfinalcopy 

\usepackage{booktabs}
\usepackage{multirow}
\usepackage{array}
\usepackage{diagbox}
\usepackage{tabularx}
\usepackage{amssymb}
\usepackage{xcolor}
\usepackage{pifont}
\usepackage{amsmath}
\usepackage{color,colortbl}
\usepackage{tabularx}
\usepackage{subcaption}
\usepackage{makecell}
\usepackage{xspace}
\usepackage{xargs}  
\usepackage{todonotes}
\usepackage{hyperref}
\usepackage{arydshln}
\usepackage{fdsymbol}
\usepackage{pgfplots}
\usepackage{pgfplotstable}
\usepackage{arydshln}

\newcommand{\greencheck}{{\ding{51}}}
\newcommand{\redcross}{{\ding{55}}}
\definecolor{intcolor}{HTML}{848FA2}
\definecolor{editcolor}{HTML}{CC2D35}
\definecolor{repcolor}{HTML}{058ED9}
\definecolor{orange}{HTML}{848FA2}
\definecolor{red}{HTML}{CC2D35}
\definecolor{teal}{HTML}{058ED9}
\newcommand{\squad}{\textsc{SQuAD}\xspace}
\newcommand{\squadone}{\textsc{SQuAD}-v1\xspace}
\newcommand{\squadtwo}{\textsc{SQuAD}-v2\xspace}
\newcommand{\corpus}{\textsc{Disfl-QA}\xspace}
\newcommand{\PreserveBackslash}[1]{\let\temp=\\#1\let\\=\temp}
\newcolumntype{C}[1]{>{\PreserveBackslash\centering}p{#1}}
\newcolumntype{R}[1]{>{\PreserveBackslash\raggedleft}p{#1}}
\newcolumntype{L}[1]{>{\PreserveBackslash\raggedright}p{#1}}
\newcolumntype{s}{>{\hsize=.5\hsize}X}
\newcolumntype{M}[1]{>{\centering\arraybackslash}m{#1}}

\title{\corpus: A Benchmark Dataset for Understanding Disfluencies\\ in Question Answering}

\author{Aditya Gupta$^\spadesuit$~~~~~~Jiacheng Xu$^\diamondsuit$${\thanks{  ~~Work done during an internship at Google.}}$~~~~~~Shyam Upadhyay$^\spadesuit$~~~~~~Diyi Yang$^{\clubsuit}$~~~~~~Manaal Faruqui$^\spadesuit$ \\
$^\spadesuit$Google Assistant \\
$^\diamondsuit$The University of Texas at Austin \\
$^{\clubsuit}$Georgia Institute of Technology \\
\texttt{disfl-qa@google.com} \\
  }

\date{}

\begin{document}
\maketitle
\begin{abstract}
\emph{Disfluencies} is an under-studied topic in NLP, even though
it is ubiquitous in human conversation. This is largely due to the lack of datasets 
containing disfluencies. In this paper, we present a new 
challenge question answering dataset, \corpus, a derivative of \textsc{SQuAD}, where humans introduce contextual disfluencies in previously fluent questions. 
\corpus contains a variety of challenging disfluencies that require a more comprehensive understanding of the text than what was necessary in prior datasets. 
Experiments show that the performance of existing state-of-the-art question answering models degrades significantly when tested on \corpus in a zero-shot setting.
We show data augmentation methods partially recover 
the loss in performance and also demonstrate the efficacy of using gold data for fine-tuning. We argue that we need large-scale disfluency datasets in order for NLP models to be robust to them. The dataset is publicly available at: \url{https://github.com/google-research-datasets/disfl-qa}.
\end{abstract}

\section{Introduction}
\label{sec:intro}

\begin{figure}[t!]
\centering
\footnotesize
\begin{subfigure}[t]{\linewidth}
\begin{tabularx}{\linewidth}{X}
\toprule
\textbf{Repetition} \hfill When is \textcolor{editcolor}{\textbf{Eas}} \textcolor{intcolor}{\textbf{ugh}} \textcolor{repcolor}{\textbf{Easter}} this year?\\
\textbf{Correction}  \hfill When is \textcolor{editcolor}{\textbf{Lent}} \textcolor{intcolor}{\textbf{I meant}} \textcolor{repcolor}{\textbf{Easter}} this year?\\
\textbf{Restarts}  \hfill \textcolor{editcolor}{\textbf{How much}} \textcolor{intcolor}{\textbf{no wait}}  \textcolor{repcolor}{\textbf{when is}} Easter this year? \\
\bottomrule
\caption{Conventional categories of \textit{Disfluencies}. The \textcolor{editcolor}{\textbf{\textit{reparandum}}} (words intended to be corrected or ignored), \textcolor{intcolor}{\textbf{\textit{interregnum}}} (optional discourse cues) and \textcolor{repcolor}{\textbf{\textit{repair}}} are marked.}
\label{fig:gen-disf}
\end{tabularx}
\end{subfigure}
\begin{subfigure}[t]{\linewidth}
\begin{tabularx}{\linewidth}{X}
\toprule

 \textbf{Passage:} {\it The Normans (Norman: Nourmands; French: Normands; Latin: Normanni) were the people who in the 10th and 11th centuries gave their name to Normandy, a region in France. They were descended from Norse ("Norman" comes from "Norseman") raiders and pirates from Denmark, Iceland and Norway who, under their leader Rollo,  \ldots}
\\
\\
$\mathbf{q_1}$: In what country is Normandy located? \\
$\mathbf{dq_1}$: In what country is \textcolor{editcolor}{\textbf{Norse}} found \textcolor{intcolor}{\textbf{no wait}} \textcolor{repcolor}{\textbf{Normandy not Norse}}?\\
$\mathbf{T5(q_1)}$: France \greencheck \\
$\mathbf{T5(dq_1)}$: Denmark \redcross \\
\vspace{0em}
$\mathbf{q_2}$: When were the Normans in Normandy?\\
$\mathbf{dq_2}$: \textcolor{editcolor}{\textbf{From which countries}} \textcolor{intcolor}{\textbf{no tell me}} \textcolor{repcolor}{\textbf{when were the Normans in Normandy}}?\\
$\mathbf{T5(q_2)}$: 10th and 11th centuries
 \greencheck \\
$\mathbf{T5(dq_2)}$: Denmark, Iceland and Norway \redcross \\

\bottomrule
\end{tabularx}
\caption{Contextualized \textit{Disfluencies} in \corpus (\S\ref{sec:annotation}).}
\label{fig:intro}
\end{subfigure}
\caption{\textbf{(a)} Categories of disfluencies~\cite{shriberg1994preliminaries} \textbf{(b)}  A passage and questions (q$_i$) from \textsc{SQuAD}, along with their disfluent versions (dq$_i$) and predictions from a \textsc{T5}-QA model.}
\end{figure}

During conversations, humans do not always premeditate exactly what they are
going to say; thus a natural conversation often includes interruptions like 
repetitions,
restarts, or corrections. 
Together these phenomena are referred to as
\textit{disfluencies}~\cite{shriberg1994preliminaries}. 
Figure~\ref{fig:gen-disf} shows
different types of conventional disfluencies in an utterance, as described by \citet{shriberg1994preliminaries}.

With the growing popularity of voice assistants, such disfluencies are of particular interest for goal-oriented or information seeking dialogue agents, because an NLU system, trained on fluent data, can easily get misled due to their presence. 
Figure~\ref{fig:intro} shows how the presence of \emph{disfluencies} in a question answering (QA) setting, namely \textsc{SQuAD} \cite{rajpurkar-etal-2018-know}, affects the prediction of a state-of-the-art T5 model~\cite{2020t5}. 
For example, the original question $q_1$ is seeking an answer
about the location of \emph{Normandy}. In the disfluent version $dq_1$ (which is \textbf{semantically equivalent} to $q_1$), the user starts asking about \emph{Norse}  and then corrects themselves to ask about the \emph{Normandy} instead. The presence of this correctional
disfluency confuses the QA model, which tend to rely on shallow textual cues from question for making predictions.

Unfortunately, research in NLP and speech community has been impeded by the lack of curated datasets containing such disfluencies. 
The datasets available today are mostly conversational in nature, and span a limited number of very specific domains (e.g., telephone conversations, court proceedings)~\cite{godfrey1992switchboard,Zayats2014MultidomainDA}. Furthermore, only a small fraction of the utterances in these datasets contain disfluencies, with a limited and skewed distribution of disfluencies types. In the most popular dataset in the literature, the \textsc{Switchboard} corpus~\cite{godfrey1992switchboard},  only $5.9\%$  of the words are \emph{disfluencies}~\cite{charniak-johnson-2001-edit}, of which $>50\%$  are \emph{repetitions}~\cite{shriberg1996disfluencies}, which has been shown to be the relatively simpler form of disfluencies~\cite{Zayats2014MultidomainDA,jamshid-lou-etal-2018-disfluency,zayats2019disfluencies}.

To fill this gap, we present \corpus, the first dataset containing 
\emph{contextual disfluencies} in an information seeking setting, namely question answering over Wikipedia passages. \corpus is constructed by asking human raters to insert disfluencies in \textit{questions} from \squadtwo, a popular question answering dataset, using the passage and remaining questions as context. These contextual disfluencies lend naturalness to \corpus, and challenge models relying on shallow matching between question and context to predict an answer. Some key properties of \corpus are: 

\begin{itemize}
    \item \corpus is a targeted dataset for disfluencies, in which all questions ($\approx$12k) contain disfluencies, making for a much larger disfluent test set than prior datasets. 
    \item Over 90\% of the disfluencies in \corpus are corrections or restarts, making it a much harder test set for disfluency correction (\S\ref{subsec:annot}).
    \item \corpus contains wider diversity in terms of semantic distractors than earlier disfluency datasets, and newer phenomenon such as coreference between the \emph{reparandum} and the \emph{repair} (\S\ref{subsec:cats}).
\end{itemize}


We experimentally reveal the brittleness of state-of-the-art LM based QA models when tested on \corpus in zero-shot setting (\S\ref{sec:zeroshot}).
Since collecting large supervision datasets containing disfluencies for training is expensive, different data augmentation methods for recovering the zero-shot performance drop are also evaluated (\S\ref{sec:heuristics}). Finally, we demonstrate the efficacy of using the human annotated data in varying fractions, for both end-to-end QA supervision and disfluency generation based data augmentation techniques (\S\ref{sec:fse}).

We argue that creation of datasets, such as \corpus, are vital for \textbf{(1)} improving understanding of disfluencies, and \textbf{(2)} developing robust NLU models in general.

\label{sec:prelim}
\begin{table*}[!t]
\small
\centering
\footnotesize
\begin{tabular}{m{1.5cm}m{6cm}m{3cm}m{3.5cm}}
\toprule
{\bf Type} & {\bf Passage} (some parts shortened) & {\bf Fluent Question} &  {\bf Disfluent Question}\\
 \midrule
 Interrogative Restart (\textbf{30\%}) & {\ldots \it Roger de Tosny travelled to the Iberian Peninsula to carve out a state for himself. In 1064, during the War of Barbastro, William of Montreuil led the papal army \ldots} & Who was in charge of the papal army in the War of Barbastro? & \textcolor{editcolor}{\textbf{Where did the}} \textcolor{intcolor}{\textbf{no}} who was in charge of the papal army in the Barbastro War? \\ 
 \midrule
 Entity Correction (\textbf{25.6\%}) & {\ldots \it While many commute to L.A. and Orange Counties, there are some differences in development, as most of San Bernardino and Riverside Counties were developed in the 1980s and 1990s\ldots } & Other than the 1980s, in which decade did most of San Bernardino and Riverside Counties develop? & Other than \textcolor{editcolor}{\textbf{the 1990s}} \textcolor{intcolor}{\textbf{I mean actually}} \textcolor{repcolor}{\textbf{the 1980s}} which decade did San Bernardino and Riverside counties develop?\\ 

 \midrule
 Adverb/Adj. Correction (\textbf{20\%}) & {\ldots \it Southern California is home to Los Angeles International Airport, the second-busiest airport in the United States by passenger volume; San Diego International Airport the busiest single runway airport in the world\ldots } & What is the second busiest airport in the United States?  & What airport in the United States is the \textcolor{editcolor}{\textbf{busiest}} \textcolor{intcolor}{\textbf{no}} \textcolor{repcolor}{\textbf{second busiest}}? \\ 
 \midrule
 Entity Type Correction (\textbf{21.1\%}) & {\ldots \it To the east is the Colorado Desert and the Colorado River, and the Mojave Desert at the border with Nevada. To the south is the Mexico-United States border\ldots } & What is the name of the water body that is found to the east? &  What is the name of the \textcolor{editcolor}{\textbf{desert}} \textcolor{intcolor}{\textbf{wait}} \textcolor{repcolor}{\textbf{the water body}} that is found to the east? \\ 
 \midrule
Others (\textbf{3.3\%}) & {\ldots \it Complexity measures are very generally defined by the Blum complexity axioms. Other complexity measures used in complexity theory include communication complexity and decision tree complexity\ldots } & What is typically used to broadly define complexity measures? & What is \textcolor{editcolor}{\textbf{defined}} \textcolor{intcolor}{\textbf{no}} is \textcolor{repcolor}{\textbf{typically used}} to broadly define complexity measures?\\ 
 \bottomrule
 \end{tabular}
 \caption{Example passage and fluent questions  from the \squad dataset and their disfluent versions provided by human raters, categorized by the type of disfluency along with their estimated percentage in the \corpus dataset.}
\label{tab:categories}
\end{table*}

\section{\corpus: Adding Disfluencies to QA}
\label{sec:annotation}
\corpus builds upon the existing \squadtwo dataset, a question answering dataset which contains curated paragraphs from Wikipedia and associated questions. Each question associated with the paragraph is sent for a human annotation task to add a contextual disfluency using the paragraph as a source of distractors. Finally, to ensure the quality of the dataset, a subsequent round of human evaluation with an option to re-annotate is conducted. 

\subsection{Source of Questions}
\label{subsec:source}
We sourced passages and questions from \squadtwo~\cite{rajpurkar-etal-2018-know} development set. \squadtwo is an extension of \squadone \cite{rajpurkar-etal-2016-squad} that contains unanswerable questions written adversarially by crowd workers to look similar to answerable ones from \squadone. We use both answerable and unanswerable questions for each passage in the annotation task. 
 
\subsection{Annotation Task} 
\label{subsec:annot}
To ensure high quality of the dataset, our annotation process consists of 2 rounds of annotation:

\paragraph{First Round of Annotation.}Expert raters were shown the passage along with all the associated questions and their answers, with one of the question-answer pair highlighted for annotation.\footnote{The raters were linguistic experts, and were trained for the task with 2 rounds of pilot annotation.} The raters were instructed to use the provided context in crafting disfluencies to make for a non-trivial dataset. 

The rater had to provide a disfluent version of the question that (a) is \emph{semantically equivalent} to the original question (b) is \emph{natural}, i.e., a human can utter them in a dialogue setting. When writing the disfluent version of a question, we instructed raters not to include partial words or filled pauses (e.g., ``\emph{um}'', ``\emph{uh}'', ``\emph{ah}'' etc.), as they can be detected relatively easily~\cite{johnson-charniak-2004-tag,jamshid-lou-johnson-2017-disfluency}. 
Raters were shown example disfluencies from each of the categories in Table~\ref{tab:categories}. On average, raters spent $2.5$ minutes per question. Introduction of a disfluency increased the mean length of a question from $10.3$ to $14.6$ words.

\paragraph{Human Evaluation + Re-annotation.} To assess and ensure high quality of the dataset, we asked a another set of human raters the following yes/no questions: 

\begin{enumerate}
      \item Is the disfluent question \emph{consistent} with respect to the fluent question? i.e., the disfluent question is semantically equivalent to the original question in that they share the same answer.
    \item Is the disfluent question \emph{natural}? Naturalness is defined in terms of human usage, grammatical errors, meaningful distractors etc.
\end{enumerate}

After the first round of annotation, we found that the second pool of raters found the disfluent questions to be consistent and natural 96.0\% and ~88.5\% of the time, with an inter-annotator agreement of 97.0\% and 93.0\%\footnote{Cohen’s $\kappa = 0.55$, indicating moderate agreement.}, respectively.
This suggests that the initial round of annotation resulted in a high quality dataset. Furthermore, for the cases identified as either inconsistent or unnatural, we conducted a second round of re-annotation with updated guidelines to make required corrections.

\subsection{Categories of Disfluencies}
\label{subsec:cats}
To assess the distribution of different types of disfluencies, we sampled $500$ questions from the training and development sets and manually annotated the nature of disfluency introduced by the raters. Table~\ref{tab:categories} shows the distribution of these categories in the dataset. 

A notable difference between \corpus and \textsc{Switchboard}~\cite{godfrey1992switchboard} is that \corpus contains a larger fraction of corrections and restarts, which have been shown to be the hardest disfluencies to detect and correct~\cite{Zayats2014MultidomainDA,jamshid-lou-etal-2018-disfluency,yang-etal-2020-planning}. From Table~\ref{tab:categories}, we can see that $\approx$30\% and $>$65\% of the disfluencies in \corpus are restarts and corrections respectively.

In addition to the specific categories mentioned in Table~\ref{tab:categories}, the dataset includes other challenging phenomena which are shared across these categories. For instance, example below shows disfluencies which introduce \emph{coreferences} between the \emph{reparandum} and the \emph{repair} (mentions marked \textbf{[.]}), allowing more complex corrections not present in existing datasets:\\\\
\noindent  {\small \begin{tabular}{M{2.2cm}M{0.2cm}M{4.1cm}}
{Who does BSkyB have an operating license from ?} & $\rightarrow$
& {\textcolor{editcolor}{\textbf{Who  removed [BSkyB's] operating license}} \textcolor{intcolor}{\textbf{no scratch that}} \textcolor{repcolor}{\textbf{who do [they] have [their] operating license from} ?}}\\\\
\end{tabular}
}
\noindent Table~\ref{tab:ds-comparison} summarizes the key differences between \corpus and the \textsc{Switchboard} dataset.

\begin{table}[t]
\centering
\footnotesize
\begin{tabular}{L{2.9cm}R{1.8cm}R{1.5cm}}
\toprule
\bf Dataset &  \bf Switchboard & \bf \corpus \\
\toprule
Domain                      &   Telephonic Conversations          &     Wikipedia Passages         \\
\midrule
Goal-oriented                           &    No    & Yes          \\
\midrule
Contextual & No & Yes \\
\midrule
Size (\# sentences)                                     & 7.9k       & 11.8k         \\
\midrule
Disfluencies                 &    20\%    & 100\%        \\
\midrule
Correction \& Restarts                 &    $<$50\%    & $>$90\%        \\
\midrule
Coreferences &    $<$1\%    & $\approx$10\%        \\
\bottomrule
\end{tabular}
\caption{Comparison of \corpus with \textsc{SwitchBoard}. \corpus is more diverse, contains harder disfluencies and new phenomenon like coreference. 
}
\label{tab:ds-comparison}
\end{table}
\section{Experimental Setup}
\label{sec:exp-setup}
\subsection{Models to Compare}
We use two different modeling approaches to answer disfluent questions in \corpus. 

\paragraph{LMs for QA.}
We use \textsc{BERT}~\cite{devlin-etal-2019-bert} and \textsc{T5}~\cite{2020t5} as our QA models in the standard setup which has shown to achieve state-of-the-art performance for \squad. We fine-tune \textsc{BERT} for a span selection task, whereby predicting \texttt{start} and \texttt{end} probabilities for all the tokens in the context.

\textsc{T5} is finetuned under the standard \emph{text2text} formulation, when given {(question, passage)} as input the model generates the {answer} as the output. For predicting \texttt{<no} \texttt{answer>}, the model was trained to generate \emph{``unknown''}.

\paragraph{LMs for Disfluency Correction.} We also fine-tune the above LMs as disfluency correction models. Given the disfluent question as input, a correction model predicts the fluent question, which is then fed into a QA model. For BERT, we use the state-of-the-art \textsc{BERT}-based disfluency correction model by \citet{jamshid-lou-johnson-2020-improving} trained on \textsc{Switchboard}. 
We also train \textsc{T5} models on \corpus to prevent the distribution skew between \textsc{Switchboard} and \corpus, and account for new phenomena like coreferences. 

\input{heuristicstable}
\subsection{Training Settings}
We train the BERT and T5 variants on the following two data configurations:
\paragraph{\textsc{ALL}} where the model is trained on all of \squadtwo, including the non-answerable questions. Evaluation is done against the entire test set.
\paragraph{\textsc{ANS}} where the model is trained only on answerable questions from \squadone, without the capabilities of handling non-answerable questions.

\input{zstable}
\subsection{Datasets}
\paragraph{Human Annotated Datasets.}
We use 3 datasets in our experiments: \squadone, \squadtwo, and \corpus. 
We split the $11,825$ annotated questions in \corpus into train/dev/test set containing $7182/1000/3643$ questions, respectively. The split was also done at an article level such that the questions belonging to the same passage belong in the same split.
For zero-shot experiments, we only use the train of \squad.  

Evaluation is done on the subset of SQuAD-v2 development set that corresponds to the \corpus test to ensure fair comparison.

\paragraph{Heuristically Generated Data.}
\label{sec:heuristics}
We also generate disfluencies heuristically to  validate  the  importance of human annotated disfluencies. Inspired by the disfluency categories seen in our annotation task, we derive the following heuristics to augment our data with \textit{silver}\footnote{The \textit{silver} nature of the data is due to the fact that we can not enforce naturalness or semantic equivalence of \S\ref{sec:annotation}.} standard disfluencies: (i) \textsc{Switch-Q} which inserts prefix of another question as a prefix to the original question, and (ii) \textsc{Switch-X}, where \textsc{X} could be verb, adjective, adverb, or entity, and is inserted as a reparandum in the question.

To facilitate contextual disfluencies, we use the reparandums from the context. For \textsc{Switch-Verb/Adj/Adv/Ent}, this was done by picking tokens and phrases from the context passage. For \textsc{Shift-Q}, we used other questions associated with the same passage.
We used spaCy\footnote{\url{https://spacy.io/}} NER and POS tagger to extract relevant entities and POS tags, and sample interregnum from a list of fillers. Table~\ref{tab:heuristics} shows an example from each of the heuristics. We then finally combine all the heuristics (\textsc{All} in Table~\ref{tab:heuristics}) by uniformly sampling a single disfluent question from the set of possible transformations of the question. 


\subsection{Evaluation Method} 
In all our experiments, we evaluate QA performance using the standard \squadtwo \href{https://worksheets.codalab.org/rest/bundles/0x6b567e1cf2e041ec80d7098f031c5c9e/contents/blob/}{evaluation script} which reports EM and F1 scores over the \texttt{HasAns} (asnwerable) and \texttt{NoAns} (non-answerable) slices along with the overall scores. For brevity, we report only the F1 numbers as we observed similar trends in EM and F1 across our experiments. 

\section{Experiments}

We conduct experiments with \corpus to answer the following questions: \textbf{(a)} Are state-of-the-art LM based QA models robust to introduction of disfluencies in the questions under a zero-shot setting ?
    \textbf{(b)} Can we use heuristically generated synthetic disfluencies to aid the training of QA models to handle disfluencies ?
    \textbf{(c)} Given a small amount of labeled data, can we recover performance by fine-tuning the QA models  or training a disfluency correction model to pre-process the disfluent questions into fluent ones before inputting to the QA models ?
    \textbf{(d)} In the above setting, can we train a generative model to generate more disfluent training data ?

\subsection{Zero-Shot Performance}
\label{sec:zeroshot}
Table~\ref{tab:zse} shows the performance of different variants measuring their zero-shot capabilities.  

\paragraph{Performance of BERT-QA and T5-QA.} We see from Table~\ref{tab:zse} that when tested directly on on heuristics and \corpus test sets, both the BERT-QA and T5-QA models exhibit significant performance drop, as compared to the performance on the fluent benchmark of \squad.  The performance drop for the complete models is greater when compared to their answerable-only counterparts. The best performing T5-ALL model shows a \textbf{drop of} $\mathbf{27.95}$ \textbf{F1} points for the complete setup and $\mathbf{13.32}$ \textbf{F1} point for the answerable only T5-ANS model. This shows BERT and T5 are not robust when questions contain disfluencies.

\paragraph{Disfluency Correction + T5-QA.} We use the \textsc{BERT} based state-of-the-art disfluency correction \cite{jamshid-lou-johnson-2020-improving} as a pre-processing step before feeding the input to our T5-QA model. The models trained on \textsc{Switchboard} are not able to fill a significant performance gap, with the complete and answerable models recovering $4.07$ and $2.25$ F1 points, respectively. We will revisit this setting in the few-shot experiments.

\paragraph{\corpus test-set vs.\ Heuristics test-set.} Next, we compare the performance of heuristically generated disfluent questions against the human annotated questions. In general, human annotated disfluent questions exhibit larger performance drop compared to heuristics, across different models. 
 
 Taking a closer look at the T5-ALL model shows that \corpus shows a bigger drop in \texttt{HasAns} cases and smaller increase in \texttt{NoAns} cases, as compared to the heuristics test set. For the T5-ANS model, \corpus shows a larger drop in performance which is attributed to the model picking wrong answer span. Based on this, we hypothesize that between the two datasets, heuristics are able to confuse the models in over-predicting \texttt{<no} \texttt{answer>}, but \corpus is superior when it comes to confuse the models to picking a different answer span altogether (as seen in Table~\ref{tab:zse} for models in ANS setting). This demonstrates that collecting a dataset like \corpus via human annotation holds value for contextual disfluencies.
 


\begin{table}[t]
\centering
\footnotesize
\begin{tabular}{L{1.7cm}C{1.2cm}C{1.3cm}C{1.2cm}}
\toprule
\bf Original & \multicolumn{2}{c}{\textbf{\texttt{HasAns}}} & \textbf{\texttt{NoAns}} \\
\midrule
\bf Prediction & \textbf{\texttt{NoAns}} & \textbf{\texttt{WrongAns}} & \textbf{\texttt{HasAns}}  \\
\midrule
\squad & $71$ & $150$ & $216$ \\
 \corpus & \textcolor{red}{$\mathbf{1091}$} & $168$ & $174$ \\
 \bottomrule
 \end{tabular}
 \caption{Breakdown of prediction errors for the T5-QA-ALL model on the fluent and disfluent questions. \texttt{WrongAns} represents that the model predicted an incorrect span from context.}
  \label{tab:predictionshift}
\end{table}

\paragraph{Performance Gap Breakdown.} For models trained on ALL setting, we find that the performance drop is largely due to the drop in F1 (over $\mathbf{50}$ points) on \texttt{HasAns} questions as opposed to \texttt{NoAns} questions, where it is almost negligible or even positive in some cases. Upon closer analysis (Table~\ref{tab:predictionshift}) we find that a major fraction of prediction errors for \texttt{HasAns} is attributed to \texttt{HasAns} $\rightarrow$ \texttt{NoAns} errors, instead of \texttt{HasAns} $\rightarrow$ \texttt{WrongAns}.\footnote{We use the standard \squad evaluation script and mark a prediction as \texttt{WrongAns} iff  F1(pred,gold)$<0.8$.}

\begin{table}[t]
\centering
\footnotesize
\begin{tabular}{L{2cm}C{1.1cm}C{1.1cm}C{1.1cm}}
\toprule
 & \bf \makecell{HasAns \\ F1} & \bf  \makecell{NoAns \\ F1} & \bf  \makecell{Overall \\ F1} \\
\toprule
Fluent $(\star)$ & $91.38$ & $87.67$ & $89.59$\\
Zero-Shot & $35.21$ & $90.06$ & $61.64$ \\
\midrule
+ \textsc{Sw-Adj} & $68.49$ & $86.24$ & $77.03$ \\
+ \textsc{Sw-Adv} & $67.37$ & $85.27$ & $75.98$ \\
+ \textsc{Sw-ENT} & $\mathbf{74.76}$ & $\mathbf{85.95}$ & $\mathbf{80.14}$\\
+ \textsc{Sw-Q} & $70.03$ & $78.94$ & $74.31$ \\
+ \textsc{Sw-Verb} & $68.01$ & $87.16$ & $77.22$ \\
\midrule
+ \textsc{All} & $\mathbf{78.86}$ & $\mathbf{85.96}$ & $\mathbf{82.27}$\\
\bottomrule
\end{tabular}
\caption{Performance on \corpus with individual (\textsc{Sw-XX}) and combined  (\textsc{All}) heuristics based data augmentation and fine-tuning. }
\label{tab:heuristics-perf}
\end{table}

We believe that the disfluencies are causing the answerable questions to resemble the non-answerable ones as seen by both BERT and T5 models under ALL setting. This results in an overly conservative model in terms of answer-ability and instead resorts to over-predicting \texttt{<no~answer>}, causing gain in non-answerable recall at the cost of precision. 
In contrast, for a comparable ANS model the drop in F1 is smaller, primarily due to relatively easier decision making, i.e. not required to decide when to answer vs. not. 

\paragraph{Fine-tuning on Heuristic Data.}
In this experiment, we fine-tune on heuristically generated data from \S\ref{sec:heuristics} and directly test on \corpus.
Table~\ref{tab:heuristics-perf} compares the performance of the heuristics fine-tuned model on the \corpus test-set. The overall heuristics trained model (\textsc{All}) is able to cover a significant performance drop from $61.64$ to $82.27$, an increase of $20.63$ F1 points. However, this still is $7.32$ F1 points short of the fluent performance. 

Amongst the individual heuristics, we observe the following order of effectiveness w.r.t. performance on the \texttt{HasAns} cases: $ \textsc{ENT}>\textsc{SQ}>\textsc{ADJ}>\textsc{VERB}>\textsc{ADV}$. One possible explanation for \textsc{Switch-ENT} and \textsc{Switch-Q} being more effective is the fact that our original annotated dataset has a relatively high percentage of entity and interrogative correction. 

\subsection{Few Shot Performance}
\label{sec:fse}
Next, we evaluate the performance of the models when we use a part of human annotated gold disfluent data for training: (i) direct end-to-end supervision, (ii) generation based data augmentation, and (iii) training disfluency correction models.

\begin{figure}[t]
  \centering
\begin{tikzpicture}[scale=0.9]
\begin{axis}[
	ylabel=F1,
	xlabel=Percentage of DISFL-QA Training Data,
	width=9cm,height=7cm,
	y label style={at={(axis description cs:0.05,.5)}},
    legend style={at={(0.8,0.05)},anchor=south}
    ]


\addplot[color=blue,mark=o] coordinates {
	(0, 35.31)
	(1, 63.33)
	(5, 77.01)
	(10, 81.21)
	(25, 83.58)
	(50, 85.09)
	(100, 86.40)
};

\addplot[color=red,mark=*] coordinates {
	(0, 90.06)
	(1, 84.42)
	(5, 82.02)
	(10, 82.53)
	(25, 83.84)
	(50, 85.33)
	(100, 86.53)
};
\legend{HasAns, NoAns}
\node[label={360:{\small (0, 35.3)}},inner sep=-1pt] at (axis cs:0,35.31) {};
\node[label={360:{\small (0, 90.1)}},inner sep=-1pt] at (axis cs:0,90.06) {};
\node[label={270:{\small (25, 83.5)}},inner sep=-1pt] at (axis cs:25,83.5) {};
\node[label={90:{\small (50, 85.2)}},inner sep=-1pt] at (axis cs:50,85.2) {};
\node[label={270:{\small (100, 86.5)}},inner sep=-1pt] at (axis cs:98,86.5) {};
\end{axis}
\end{tikzpicture}
\caption{Few shot performance for different fraction of training data. We can see that performance on \texttt{HasAns} cases increases monotonically with increase in gold data. However, for the \texttt{NoAns} cases, the performance first takes a drop (compared to zero-shot) and then increases. }
\label{fig:k-shot}
\end{figure}
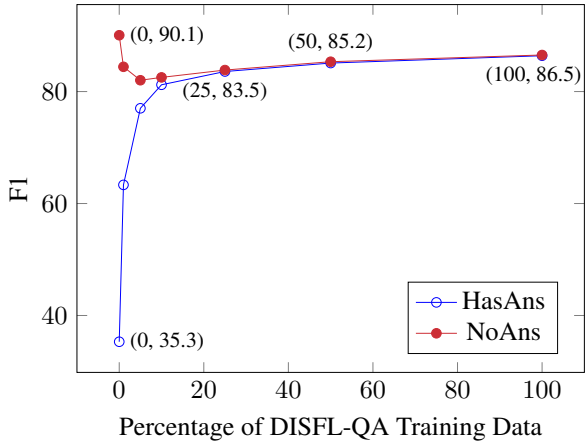

\paragraph{Direct Supervision (\emph{k}-shot).}
\label{sec:direct}
In this setting, we pick a \squadtwo T5 model and then perform a second round of fine-tuning with varying percentages of \corpus gold training data. We experiment with $1, 5, 10, 25, 50,$ and $100$ percent of the total gold data. 

Figure~\ref{fig:k-shot} shows the performance for the \texttt{HasAns} and \texttt{NoAns} cases as we increase the amount of training data. The \texttt{HasAns} performance increases gradually from $35.31$ F1 points, in the zero-shot setting, to $86.40$ F1 points with complete training data. Interestingly, for the \texttt{NoAns} cases, the performance first drops from $90.06$ F1 points, in the zero-shot setting, to $82.02$ F1 with $5\%$ data and then monotonically increasing to $86.53$ F1 with complete data. This can be attributed to the fact that the zero-shot models were under-predictive (high recall, low precision for \texttt{<no asnwer>}) due to lack of robustness to disfluent inputs.

Furthermore, Table~\ref{tab:fse-perf} compares the performance of using the gold training data of \corpus against the heuristics data. It shows that the models trained with disfluent data from \corpus are able to cover a major gap in answerable slice, which wasn't possible with the heuristically generated data. Direct supervision bring an additional performance improvement of $4.19$ F1 points over the heuristics.
\begin{table}[t]
\centering
\footnotesize
\begin{tabular}{L{2.4cm}C{1.1cm}C{1.1cm}C{1.1cm}}
\toprule
 & \bf \makecell{HasAns \\ F1} & \bf  \makecell{NoAns \\ F1} & \bf  \makecell{Overall \\ F1} \\
\toprule
Fluent $(\star)$ & $91.38$ & $87.67$ & $89.59$\\
Zero-Shot & $35.21$ & $90.06$ & $61.64$ \\
Heuristics & $78.86$ & ${85.96}$ & ${82.27}$\\
\midrule
 \multicolumn{4}{c}{Direct Supervision} \\
\midrule
$25\%$ Data & $83.58$  & $83.84$ & $83.71$ \\
~+ Q~$\rightarrow$~DQ  & $86.44$ & $84.53$  &  $85.52$ \\
~+ CQ~$\rightarrow$~DQ &  $87.47$ &  $83.11$ &  $85.37$ \\
\addlinespace[.2em]
\hdashline
\addlinespace[.2em]
$50\%$ Data & $85.09$  & $85.33$ & $85.20$ \\
\addlinespace[.2em]
\hdashline
\addlinespace[.2em]
$100\%$ Data & $86.40$  & $\mathbf{86.53}$ & $\mathbf{86.46}$ \\
~+ Q~$\rightarrow$~DQ   & $86.95$ & $85.73$  &  $86.33$ \\
~+ CQ~$\rightarrow$~DQ  &  $\mathbf{87.29}$ &  $85.22$ &  $86.29$ \\
\midrule
 \multicolumn{4}{c}{Pipelined} \\
\midrule
DQ~$\rightarrow$~Q    &  $87.65$ & $\mathbf{86.70}$  & $\mathbf{87.19}$  \\
CDQ~$\rightarrow$~Q    &  $\mathbf{87.99}$ & $86.02$  & $87.04$  \\
\bottomrule
\end{tabular}
\caption{Performance on the test set of \corpus when using gold human annotated data in training different components. }
\label{tab:fse-perf}
\end{table}

\begin{table*}[t!]
\centering
\footnotesize
\begin{tabular}{m{15.5cm}}
\toprule
\textbf{Passage}:  {\ldots \it Whereas a \textbf{genome sequence} lists the order of every DNA base in a genome, a \textbf{genome map} identifies the landmarks. A genome map is less detailed than a genome sequence and aids in navigating around the genome   \ldots}  \\ 
\\
\textbf{Fluent Question} : What does a genome map list the order of ? \\
\textbf{T5 Q~$\rightarrow$~DQ} : \textcolor{orange}{What is} no what does a genome map list the order of ? \\
\textbf{T5 CQ~$\rightarrow$~DQ} : What does a \textcolor{teal}{genome sequence} list the order of no sorry what does a genome map list the order of? \\
\midrule
\textbf{Passage}:  {\ldots \it The presence of \textbf{fat} in the small intestine produces hormones that stimulate the release of pancreatic lipase from the pancreas and \textbf{bile} from the liver which helps in \ldots}  \\ 
\\
\textbf{Fluent Question} : What is one molecule of fat ? \\
\textbf{T5 Q~$\rightarrow$~DQ} : What is one molecule of \textcolor{orange}{protein} no fat ? \\
\textbf{T5 CQ~$\rightarrow$~DQ} : What is one molecule of \textcolor{teal}{bile} no wait fat ? \\
\midrule
\textbf{Passage}:  {\ldots \it In 1964, \textbf{Nikita Khrushchev} was removed from his position of power and replaced with \textbf{Leonid Brezhnev}. Under his rule, the Russian SFSR \ldots}  \\ 
\\
\textbf{Fluent Question} : When did Leonid Brezhnev die ?  \\
\textbf{T5 Q~$\rightarrow$~DQ} : \textcolor{red}{When was the age of Leonid Brezhnev ?}  \\
\textbf{T5 CQ~$\rightarrow$~DQ} : When did \textcolor{teal}{Nikita Khrushchev} \emph{er} I mean Leonid Brezhnev die ? \\
 \bottomrule
 \end{tabular}
 \caption{Example disfluent question (DQ) as generated by the Q~$\rightarrow$~DQ and CQ~$\rightarrow$~DQ T5 generative models for data augmentation. We observe that  CQ~$\rightarrow$~DQ generates \textcolor{teal}{meaningful} disfluencies compared to context-free generation, the latter leading to \textcolor{orange}{irrelevant} or \textcolor{red}{inconsistent} questions in some cases.}
\label{tab:generation}
\end{table*}

\paragraph{Generation Based Data Augmentation.} 
We use the \textsc{T5} model for synthetically generating disfluent question from fluent question in the \emph{text2text} framework.
We use the training set of \corpus to train the following generative models: (i) context-free generation (Q~$\rightarrow$~DQ), and (ii) context-dependent generation (CQ~$\rightarrow$~DQ) which use passage as well for generation. 

Table~\ref{tab:generation} shows example generation from the two models. We observe that CQ~$\rightarrow$~DQ is able to learn meaningful contextual disfluency generation, whereas Q~$\rightarrow$~DQ can lead to non-meaningful or inconsistent disfluencies due to lack to context.

We then pick $5$k random (question, answer) pairs from \squad training data and apply our generative model to produce disfluent training data for the QA models. Table~\ref{tab:fse-perf} shows the performance of using data augmentation. 
We perform data augmentation under two different train data settings: (1) $25\%$ data, and (2) $100\%$ data. Interestingly, for the models trained on $25\%$ train data + generated data, we observe a gain of $1.81$ F1 points ($83.71\rightarrow85.52$) in the overall performance which is close to the absolute performance of using $50\%$ gold data. However, for the setup with $100\%$ gold data + generated data, we did not observe a similar improvement in the overall performance.

\paragraph{Pipelined: Disfluency Correction + QA.} 
Unfortunately, existing disfluency correction models and datasets assume that fluent text is a sub-sequence of the disfluent one, and hence these approaches cannot solve disfluencies in \corpus involving coreference. For fair comparison, we train a T5 generation model as a \corpus specific disfluency correction model using the training set of \corpus, with a simple  DQ~$\rightarrow$~Q and CDQ~$\rightarrow$~Q \textsc{T5} task formulation. 

With this pipelined approach, we get further improvements with an overall F1 of $87.19$ (Table~\ref{tab:fse-perf}), however, still lacking by $\approx$2.4 F1 points compared to the fluent dataset. This shows that such complex cases require better modeling, preferably in an end-to-end setup.
\label{sec:pipeline}

\pgfplotstableread[col sep=comma]{ 
    frac, time_ws, time_full
    10, 85.99, 86.48
    20, 13.17, 56.33
    30, 12.39, 56.40
    40, 12.18, 55.83
    50, 11.95, 56.40
    60, 11.82, 56.56
    70, 11.71, 56.18
    80, 11.01, 56.65
    90, 11.15, 56.47
    100, 90.84, 96.71
  }\speedupdata

\section{Related Work}

\subsection{Disfluency Correction}
The most popular approach in literature poses disfluency correction as a sequence tagging task, in which the fluent version of the utterance is obtained by identifying and removing the disfluent segments~\cite{Zayats2014MultidomainDA,ferguson-etal-2015-disfluency,Zayats2016DisfluencyDU,lou2017disfluency,jamshid-lou-johnson-2020-improving,wang2020multi}. .
Traditional disfluency correction models use syntactic features~\cite{Honnibal14a}, language models~\cite{Johnson2004AnIM,zwarts-johnson-2011-impact}, discourse markers~\cite{Crible2017DiscourseMA}, or prosody-based features for learning~\cite{zayats-ostendorf-2019-giving,wang2017transition} while recent disfluency correction models largely utilize pre-trained neural representations~\cite{lou2018disfluency}.
Most of these models depend on human-annotated data. As a result, recently, data augmentation techniques have been proposed ~\cite{yang-etal-2020-planning, mcdougall2017profiling} to alleviate the strong dependence on labeled data. 
However,  the resulting augmented data either via heuristics \cite{wang2020multi} or generation models \cite{yang-etal-2020-planning} is often limited in terms of disfluencies types and may not well capture natural disfluencies in daily conversations. 


\subsection{Question Answering Under Noise}
In the QA literature, our work is related to two threads that aim to improve robustness of QA models: (i) QA under adversarial noise, and (ii) noise arising from speech phenomena.

Prior work on adversarial QA have predominantly generated adversaries automatically \cite{zhao2018generating}, which are verified by humans to ensure semantic equivalence (i.e. answer remains same after perturbation). 
For instance, \newcite{ribeiro-etal-2018-semantically} generated adversaries using paraphrasing, while
\citet{mudrakarta2018did} perturbed questions based on attribution. 
Closest work to ours is \newcite{jia-liang-2017-adversarial}, who modified \squad to contain automatically generated adversarial sentence insertions.

Our work is more closely related to prior work on making NLP models robust to noise arising from speech phenomena.  Earlier work~\cite{surdeanu2006design,leuski-etal-2006-building} have built QA models which are robust to disfluency-like phenomenon, but they were limited in the corpus complexity, domain, and scale. Recently there has been renewed interest in constructing audio enriched versions of existing NLP datasets, for example, the \textsc{Spoken-SQuAD} \cite{Li2018SpokenSA} and \textsc{Spoken-CoQA} \cite{you2021data} with the aim to show the effect of speech recognition errors on QA task. However, since collecting audio is challenging, another line of work involves testing the robustness of NLP models to ASR errors in transcribed texts containing synthetic noise using TTS $\rightarrow$ ASR technique \cite{Peskov2019,  peng2020raddle, liu2020robustness, ravichander2021noiseqa}. 
Our work suggests a complementary approach to data collection to surface a specific speech phenomenon that affects NLP. 


\section{Conclusion}

This work presented \corpus, a new challenge set containing contextual semantic disfluencies in a QA setting. \corpus contains diverse set of disfluencies rooted in context, particularly a large fraction of corrections and restarts, unlike prior datasets.
\corpus allows one to directly quantify the effect of presence of disfluencies in a downstream task, namely QA. We analyze the performance of models under varying when subjected to disfluencies under varying degree of gold supervision: zero-shot, heuristics, and \emph{k}-shot.
\paragraph{Large-scale LMs are not robust to disfluencies.} Our experiments showed that the state-of-the-art pre-trained models (BERT and T5) are not robust when directly tested on disfluent input from \corpus. Although a naturally occurring phenomenon, the noise introduced by the disfluent transformation led to a non-answerable behavior at large.

\paragraph{Contextual heuristics partially recover performance.} We derived heuristics, in attempt to resemble the contextual nature of \corpus, by introducing semantic distractors based on NER, POS, and other questions. In our experiments, we found that heuristics are effective in: (1) confusing the models in zero-shot setup, and (2) partially recovering the performance drop on \corpus with fine-tuning. This indicates that the heuristics might be capturing some key aspects of \corpus.

\paragraph{Efficacy of gold training data.} We use the gold data for supervising various models: (i) end-to-end QA model, (ii) disfluency correction, and (iii) disfluency generation (for data augmentation). For all the experiments, gold supervision outperforms heurisitics' supervision significantly. Furthermore, we observed that in a low resource setup generation based data augmentation can match the performance of a high resource modeling setup. \\ 
\vspace{-0.5em}
\section{Discussion}

While \corpus aims to fill a major gap between speech and NLP research community, understanding \emph{disfluencies} holistically requires the following:

\paragraph{General disfluencies focused NLP research.} We believe understanding of disfluencies is a key ingredient for enabling natural human-machine communication in the near future, and call upon the NLP community to devise generalized few-shot or zero-shot approaches to effectively handle disfluencies present in input to NLP models, without requiring task specific disfluency datasets. 

\paragraph{Constructing datasets for spoken problems.} We would also like to bring attention to the fact that being a speech phenomenon, a spoken setup would have been an ideal choice for disfluencies dataset. This would have accounted for higher degree of confusion, hesitations, corrections, etc. while recalling parts of context on the fly, which otherwise one may find hard to create synthetically when given enough time to think. 

However, such a spoken setup is extremely tedious for data collection mainly due to: (i) privacy concerns with acquiring speech data from real world speech transcriptions, (ii) creating scenarios for simulated environment is a challenging task, and (iii) relatively low yield for cases containing disfluencies. In such cases, we believe that a targeted and purely textual mode of data collection can be more effective both in terms of cost and specificity. 


\bibliographystyle{acl_natbib}
\bibliography{acl2021}


\end{document}

%% file: heuristicstable.tex
\begin{table}[t]
\centering
\scriptsize
\begin{tabular}{C{.5cm}L{2.1cm}L{3.5cm}}
\toprule
{\bf Rule} & {\bf Fluent} & {\bf Disfluent}\\
 \midrule
 \textsc{Q} & What was the Norman religion? & What was \textcolor{editcolor}{\textbf{replaced with}} \textcolor{intcolor}{\textbf{no no}} what was the Norman religion?\\ 
 \midrule
\textsc{V} &  When was the Duchy of Normandy founded? & 
When was the Duchy of Normandy \textcolor{editcolor}{\textbf{offered}} \textcolor{intcolor}{\textbf{ugh I mean}} \textcolor{repcolor}{\textbf{founded}}?\\ 
 \midrule
 \textsc{Adj} & What is the original meaning of the word Norman? & What is the \textcolor{editcolor}{\textbf{English}} \textcolor{intcolor}{\textbf{rather}} \textcolor{repcolor}{\textbf{original}} meaning of the word Norman?
 \\ 
 \midrule
  \textsc{Adv} & Who did Beyoncé perform privately for in 2011? & Who did Beyoncé perform \textcolor{editcolor}{\textbf{publicly}} \textcolor{intcolor}{\textbf{oops}} \textcolor{repcolor}{\textbf{privately}} for in 2011?
 \\ 
 \midrule
 \textsc{Ent} & Who was a prominent Huguenot in Holland? &  Who was a prominent \textcolor{editcolor}{\textbf{Saint Nicholas}} \textcolor{intcolor}{\textbf{no I mean}} \textcolor{repcolor}{Huguenot} in Holland?\\ 
 \bottomrule
 \end{tabular}
 \caption{Example of synthetically generated disfluent questions using the contextual heuristics.}
\label{tab:heuristics}
\end{table}

%% file: zstable.tex
\begin{table*}[t]
\small
\centering
\begin{tabular}{C{2cm}C{2cm}C{2cm}L{2cm}L{2cm}L{2cm}}
\toprule
\bf Model & \bf Train & \bf Eval  & \bf ~HasAns-F1 & \bf ~NoAns-F1 & \bf ~Overall-F1\\
\midrule
\multirow{6}{*}{\bf BERT-QA} & \multirow{3}{*}{\textbf{ALL}} & \textsc{SQuAD} & $83.87$ & $70.55$ & $77.46$ \\
&  & Heuristics  & $51.45$ {\scriptsize$\color{red}\downarrow32.42$} &  $74.49$ {\scriptsize$\color{teal}\uparrow3.94$} & $62.53$ {\scriptsize$\color{red}\downarrow14.93$} \\
&  &  \corpus &  $40.97$ {\scriptsize$\color{red}\downarrow42.90$} &  $75.97$ {\scriptsize$\color{teal}\uparrow5.42$} & $57.81$ {\scriptsize$\color{red}\downarrow19.65$} \\
\cline{2-6}  \\ [-1.5ex]
& \multirow{3}{*}{\textbf{ANS}} & \textsc{SQuAD} &  $89.63$ & ~~~~~~~~~~- &  $89.63$ \\
&  & Heuristics  & $80.52$ {\scriptsize$\color{red}\downarrow9.11$} & ~~~~~~~~~~-  & $80.52$ {\scriptsize$\color{red}\downarrow9.11$} \\
&  & \corpus & $78.88$  {\scriptsize$\color{red}\downarrow10.75$} & ~~~~~~~~~~- & $78.88$ {\scriptsize$\color{red}\downarrow10.75$} \\
\midrule
\multirow{6}{*}{\bf T5-QA} & \multirow{3}{*}{\textbf{ALL}} & \textsc{SQuAD} &  $91.38$ & $87.67$ &  $89.59$ \\
&  & Heuristics  &  $39.98$ {\scriptsize$\color{red}\downarrow51.40$} &  $92.57$ {\scriptsize$\color{teal}\uparrow4.90$} &  $65.27$ {\scriptsize$\color{red}\downarrow24.32$} \\
&  &  \corpus &  $35.31$ {\scriptsize$\color{red}\downarrow56.07$} &  $90.06$ {\scriptsize$\color{teal}\uparrow2.39$}  &  $61.64$ {\scriptsize$\color{red}\downarrow27.95$} \\
\cline{2-6}  \\ [-1.5ex]
& \multirow{3}{*}{\textbf{ANS}} & \textsc{SQuAD} &  $93.71$ & ~~~~~~~~~~- &  $93.71$ \\
&  & Heuristics  &  $81.73$ {\scriptsize$\color{red}\downarrow12.01$} &  ~~~~~~~~~~- &  $81.73$ {\scriptsize$\color{red}\downarrow12.01$} \\
&  & \corpus &  $80.39$ {\scriptsize$\color{red}\downarrow13.32$} &  ~~~~~~~~~~- &  $80.39$ {\scriptsize$\color{red}\downarrow13.32$} \\
\midrule
\multirow{6}{*}{\shortstack[c]{\bf Disfluency \\ \bf Correction \\ \bf + \\ \bf T5-QA} } & \multirow{3}{*}{\textbf{ALL}} & \textsc{SQuAD} & $91.38$ & $87.67$ &  $89.59$ \\
&  & Heuristics  &  $42.83$ {\scriptsize$\color{red}\downarrow48.55$}  &  $92.18$ {\scriptsize$\color{teal}\uparrow4.51$}  &  $66.56$ {\scriptsize$\color{red}\downarrow23.03$} \\
&  &  \corpus & $43.61$ {\scriptsize$\color{red}\downarrow47.77$} & $89.55$ {\scriptsize$\color{teal}\uparrow1.88$} & $65.71$ {\scriptsize$\color{red}\downarrow23.88$}  \\
\cline{2-6}  \\ [-1.5ex]
& \multirow{3}{*}{\textbf{ANS}} & \textsc{SQuAD} &  $93.71$ & ~~~~~~~~~~- &  $93.71$ \\
&  & Heuristics  & $82.27$ {\scriptsize$\color{red}\downarrow10.44$} & ~~~~~~~~~~- & $82.27$ {\scriptsize$\color{red}\downarrow10.44$} \\
&  & \corpus & $82.64$ {\scriptsize$\color{red}\downarrow11.07$} & ~~~~~~~~~~- & $82.64$ {\scriptsize$\color{red}\downarrow11.07$} \\
\bottomrule
\end{tabular}
\caption{Breakdown of zero-shot performance of fine-tuned BERT and T5 QA models, trained only on the \squad dataset, and evaluated on \squad, Heuristics (\S\ref{sec:heuristics}), and \corpus test sets. We also evaluate the performance by using state-of-the-art disfluency detection model by \citet{jamshid-lou-johnson-2020-improving} in a pipelined fashion.}
\label{tab:zse}
\end{table*}